%% file: neurips_2023.tex
\documentclass{article}
\input{1_packages}

\input{2_acronyms}

\PassOptionsToPackage{numbers,sort,compress}{natbib}  
\usepackage[final]{neurips_2023}  

\title{Adaptive Linear Embedding for Nonstationary High-Dimensional Optimization}

\author{
  Yuejiang Wen, Paul D. Franzon\thanks{Manuscript received XX YY, ZZ; revised XX YY, ZZ. Date of publication XX YY, ZZ; date of current version XX YY, ZZ. This work was partially supported by Integrated Device Technology and the National Science Foundation under Grant No. CNS 16-244770 - Center for Advanced Electronics through Machine Learning (CAEML). \textit{(Corresponding author: )}} \\
  Department of Electrical and Computer Engineering\\
  NC State University, Raleigh, NC, 27695 USA \\
  \texttt{wyuejia@ncsu.edu} \\
}

\begin{document}
\pagenumbering{arabic}

\maketitle

\begin{abstract}
  \ac{BO} in high-dimensional spaces remains fundamentally limited by the curse of dimensionality and the rigidity of global low-dimensional assumptions. While \ac{REMBO} mitigates this via linear projections into low-dimensional subspaces, it typically assumes a single global embedding and a stationary objective. In this work, we introduce Self-Adaptive cross embedding REMBO (SA-cREMBO), a novel framework that generalizes REMBO to support multiple random Gaussian embeddings, each capturing a different local subspace structure of the high-dimensional objective. An index variable governs the embedding choice and is jointly modeled with the latent optimization variable via a product kernel in a Gaussian Process surrogate. This enables the optimizer to adaptively select embeddings conditioned on location, effectively capturing locally varying effective dimensionality, nonstationarity, and heteroscedasticity in the objective landscape. We theoretically analyze the expressiveness and stability of the index-conditioned product kernel and empirically demonstrate the advantage of our method across synthetic and real-world high-dimensional benchmarks, where traditional REMBO and other low-rank BO methods fail. Our results establish SA-cREMBO as a powerful and flexible extension for scalable BO in complex, structured design spaces.
\end{abstract}

\section{Introduction}
\ac{BO} is a useful method in many ubiquitous problems such as circuit design optimization \cite{wang2020multi, torun2018global, lyu2018batch}, machine learning\cite{bergstra2013hyperopt, snoek2012practical}, computer graphics and visual design\cite{brochu2010bayesian}. It has shown promise in optimizing expensive black-box functions in low-dimensional spaces. However, its applicability to high-dimensional problems remains limited due to the exponential growth of sample complexity with input dimensionality. \ac{REMBO}~\cite{wang2013bayesian} addresses this by assuming the function of interest lies in a low-dimensional subspace and performing optimization in a projected space defined by a random linear embedding matrix $A \in \mathbb{R}^{D \times d}$, where $d \ll D$.

Despite its theoretical foundations, the performance of REMBO is highly sensitive to the choice of embedding matrix. A single random embedding may fail to intersect with the effective subspace of the objective function. To mitigate this, \emph{ensemble REMBO} methods~\cite{eriksson2019scaling, binois2020practical} explore multiple embeddings in parallel. However, training a separate Gaussian Process (GP) surrogate for each embedding leads to high computational cost, fragmented information, and reduced sample efficiency. \cite{lu2024high} introduces an additional expansion projection method in addition to the existing projection. However, the 

In this chapter, we propose a unified framework that enables multiple embeddings to be modeled jointly within a single GP surrogate. By introducing an additional discrete input dimension representing the embedding index, the surrogate model can share information across embeddings while retaining their individual characteristics. This design significantly improves model efficiency, adaptability, and generalization.

\subsection{\ac{REMBO}}
\label{sec:rembo}
\ac{REMBO} is a Bayesian optimization algorithm that uses random embeddings to transform the input space into a lower-dimensional space. The key idea behind \ac{REMBO} is that embedding the input space in a lower-dimensional space can reduce the number of function evaluations required to find the optimal solution, especially for high-dimensional problems.

The theoretical foundation of \ac{REMBO} is based on the idea of using random embeddings to preserve the pairwise distances between points in the input space. Specifically, \ac{REMBO} uses a random linear projection to embed the input space into a lower-dimensional space, such that the distances between points in the embedded space approximate the distances between points in the original space. The embedding is designed to preserve the relative distances between points, while reducing the dimensionality of the problem.

Once the input space has been embedded, \ac{REMBO} uses Bayesian optimization to search for the optimal solution in the lower-dimensional space. The algorithm maintains a probabilistic model of the objective function, which is updated after each evaluation. The model is then used to generate a set of candidate points, which are evaluated to obtain new function values.

The key advantage of \ac{REMBO} over other optimization algorithms is that it can efficiently search high-dimensional input spaces with fewer function evaluations, thanks to the embedding. The embedding reduces the curse of dimensionality, making it possible to search the space more efficiently. Additionally, the probabilistic model of the objective function enables \ac{REMBO} to balance exploration and exploitation, allowing it to quickly identify promising regions in the search space.

Although Random Embedding BO is a simple method to optimize high dimensional problems, its efficiency is low due to both embedding distortion and the over-exploration of boundary issues. Since in high dimensions data points typically lie mostly on the boundary, and anyways far away from each other, the predictive variance tends to be higher in the regions near the boundary. This is a waste of computation effort assuming the global minimum is not on the boundary\cite{malu2021bayesian}. 
Let \( f: \mathbb{R}^D \rightarrow \mathbb{R} \) be the objective function in a high-dimensional space \( D \). The goal is to find
\begin{equation}
\end{equation}

\begin{equation}
    \boldsymbol{x}^* = \arg \min_{\boldsymbol{x} \in \mathcal{X}} f(\boldsymbol{x})
\end{equation}

where \( \mathcal{X} \subset \mathbb{R}^D \) is the feasible set.

In \ac{REMBO}, the high-dimensional vector \( \mathbf{x} \) is expressed as \( \mathbf{x} = \mathbf{A}\mathbf{z} \), where \( \mathbf{z} \in \mathbb{R}^d \) (with \( d \ll D \)) is a lower-dimensional representation, and \( \mathbf{A} \) is a \( D \times d \) random matrix used to embed \( \mathbf{z} \) into the higher-dimensional space.

The optimization problem then becomes
\begin{equation}
   \boldsymbol{z}^* = \arg \min_{\boldsymbol{z} \in \mathcal{Z}} f(\boldsymbol{A}\boldsymbol{z})
\end{equation}

where \( \mathcal{Z} = \boldsymbol{A}^{-1}(\mathcal{X}) \) is the projection of the feasible set \( \mathcal{X} \) into the lower-dimensional subspace.

\section{Theoretical Foundations of REMBO}\label{sec:REMBO}
\ac{REMBO} leverages the fact that many high-dimensional optimization problems have an \textit{effective low-dimensional structure}. While the true dimension of the search space is high, the objective function \( f \) often depends on only a few key directions. REMBO exploits this property by searching in a randomly projected low-dimensional subspace, where the optimization process is computationally feasible.

However, this random projection introduces challenges that require careful consideration. Two important theoretical results are crucial in understanding the behavior of REMBO. These theorems provide guarantees about the effectiveness of random projections in preserving the structure of the function within a low-dimensional subspace, and the ability to optimize in that space while ensuring that the global optimum is not missed.

\subsection{Theorem 1: Effective Dimensionality and Decomposition}\label{sec:Theorem 1: Effective Dimensionality and Decomposition}
Suppose we want to optimize a function \( f : \mathbb{R}^D \rightarrow \mathbb{R} \) with effective dimension \( d_e \leq d \) subject to the box constraint \( X \subset \mathbb{R}^D \), where \( X \) is centered around 0. Suppose further that the effective subspace \( T \) of \( f \) is such that \( T \) is the span of \( d_e \) basis vectors, and let \( x^*_\top \in T \cap X \) be an optimizer of \( f \) inside \( T \). If \( A \) is a \( D \times d \) random matrix with independent standard Gaussian entries, there exists an optimizer \( y^* \in \mathbb{R}^d \) such that \( f(Ay^*) = f(x^*_\top) \) and \( \|y^*\|_2 \leq \sqrt{d_e} \epsilon \|x^*_\top\|_2 \) with probability at least \( 1 - \epsilon \).

\textbf{Theorem 1.}  \cite{wang2013bayesian}

Let \( f: \mathbb{R}^D \to \mathbb{R} \) be a function with effective dimensionality \( d_e \). There exists a subspace \( T \subset \mathbb{R}^D \), such that \( \text{rank}(T) = d_e \). For any \( \bm{x} \in \mathbb{R}^D \), we can decompose \( \bm{x} \) as \( \bm{x} = \bm{x}_{\top} + \bm{x}_{\perp} \), where \( \bm{x}_{\top} \in T \) and \( \bm{x}_{\perp} \in T^\perp \). Furthermore, \( f(\bm{x}) = f(\bm{x}_{\top}) \), meaning that the objective function depends only on the projection of \( \bm{x} \) onto \( T \).

The proof of this theorem shows that without loss of generality, we can restrict the optimization process to the subspace \( T \), while ensuring that the global optimum can still be found by searching within this subspace.

This theorem establishes that, due to the effective dimensionality \( d_e \) of the objective function \( f \), it is sufficient to optimize over a lower-dimensional subspace \( T \subset \mathbb{R}^D \). The proof shows that \( f \) can be decomposed into components within \( T \) and its orthogonal complement \( T^\perp \), and that \( f \) depends only on the component in \( T \). This guarantees that the function’s behavior can be captured within the low-dimensional space, thus justifying the use of random projections.

\subsection{Theorem 2: Rank Preservation and Random Projections}\label{sec:Theorem 2: Rank Preservation and Random Projections}

\textbf{Theorem 2.}  \cite{wang2013bayesian}
Let \( \bm{A} \in \mathbb{R}^{D \times d} \) be a random matrix whose columns are i.i.d. samples from \( \mathcal{N}(0, \bm{I}) \). With probability 1, the matrix \( \Phi^\top \bm{A} \) has rank \( d_e \), where \( \Phi \) is an orthonormal basis for the effective subspace \( T \). Therefore, there exists a \( \bm{y}^* \in \mathbb{R}^d \) such that the random projection \( \bm{A} \bm{y}^* \) captures the optimal point \( \bm{x}^*_\top \) in the subspace \( T \).

The proof of this theorem uses properties of Gaussian random matrices and the fact that the set of singular matrices has Lebesgue measure zero. Consequently, the rank preservation property ensures that random projections almost surely maintain the structure needed for effective optimization in the low-dimensional subspace.

This theorem ensures that random projections using a Gaussian matrix preserve the low-dimensional structure necessary for optimization. Specifically, this theorem guarantees that, with high probability, the rank of the matrix that defines the low-dimensional subspace is preserved after projection, meaning the optimizer can still be found within the subspace defined by the random projection.
\subsection{Practical Implications of These Theorems}

These two theorems together form the theoretical backbone of REMBO. \textbf{Theorem 1} ensures that optimization in a low-dimensional subspace is sufficient to capture the global optimum, while \textbf{Theorem 2} guarantees that random projections do not distort the low-dimensional subspace, thereby preserving the ability to find the optimizer. In practice, this means that REMBO can efficiently search in high-dimensional spaces by projecting the problem into a much lower-dimensional space, where Bayesian Optimization can be applied effectively. Furthermore, the rank preservation guarantees ensure that the low-dimensional projection retains the necessary geometric properties, making the method robust and scalable.

\section{Related Work}

Our method draws inspiration from the following lines of research:

\begin{itemize}
    \item \textbf{REMBO with Multiple Embeddings.} Eriksson et al.~\cite{eriksson2019scaling} and Binois et al.~\cite{binois2020practical} proposed using multiple random embedding matrices in parallel to improve robustness. However, each embedding was modeled independently, resulting in high computational cost and no information sharing.
    
    \item \textbf{Mixed-Embedding BO.} Daxberger et al.~\cite{daxberger2020mixed} introduced a joint modeling approach using a GP with a discrete embedding identifier, which is conceptually aligned with our method. Our approach formalizes this within REMBO and analyzes its implications in high-dimensional optimization.

    \item \textbf{Multi-Task BO and Structured Kernels.} Previous work in multi-task GP models~\cite{swersky2013multi} and structured kernel learning motivates our use of discrete latent inputs (e.g., embedding index) and product kernels to capture correlations across tasks or configurations.
\end{itemize}

\section{Method Description}
\begin{figure}[t]  \centering
\includegraphics[width=5in]{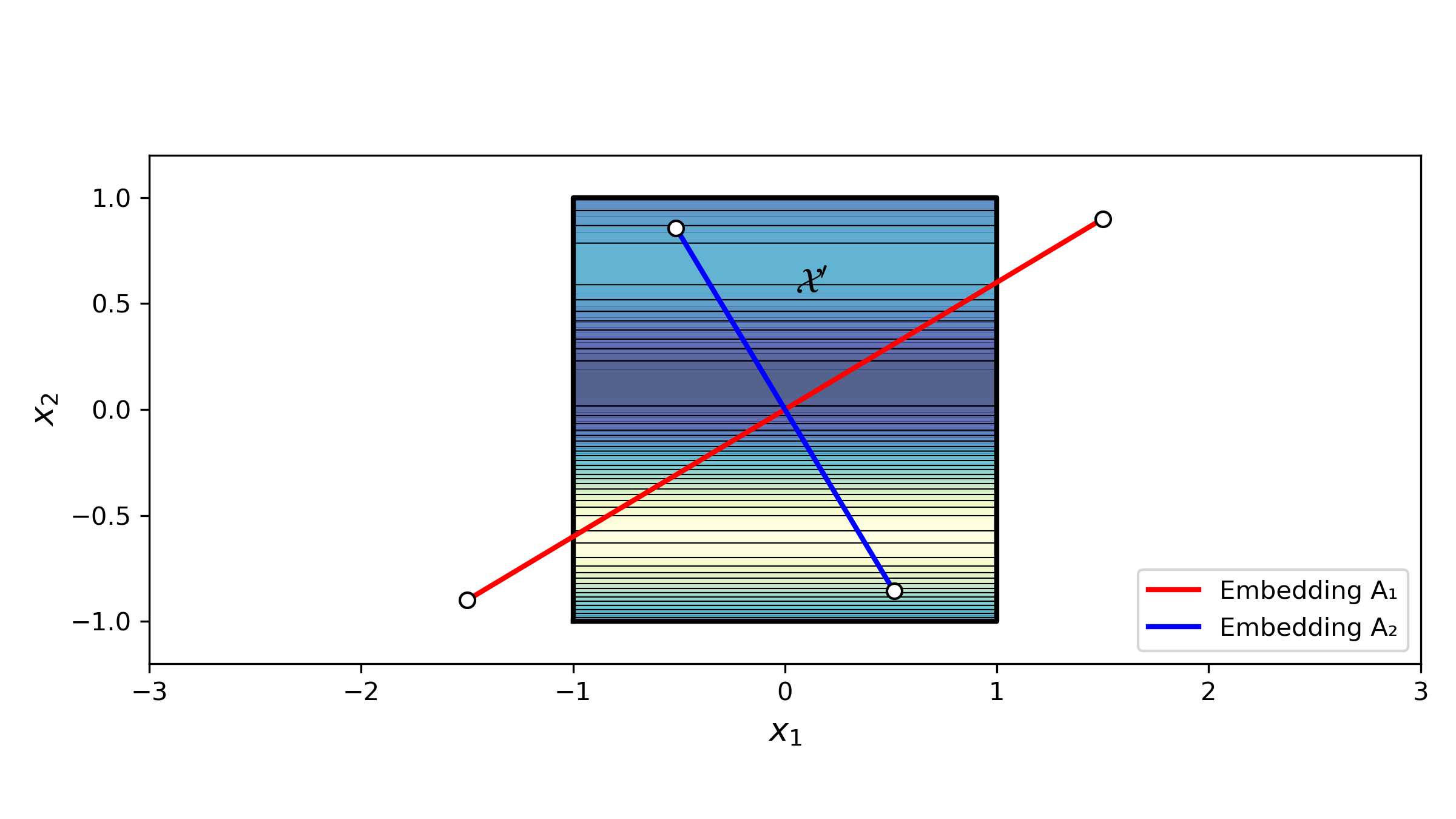}
\caption{Rndom Cross Embedding used by \ac{cREMBO} in 2D}
\label{show 2d cross embedding}
\end{figure}

\begin{figure}[t]
  \centering
  \begin{minipage}[t]{0.32\textwidth}
    \centering
    \includegraphics[width=\linewidth]{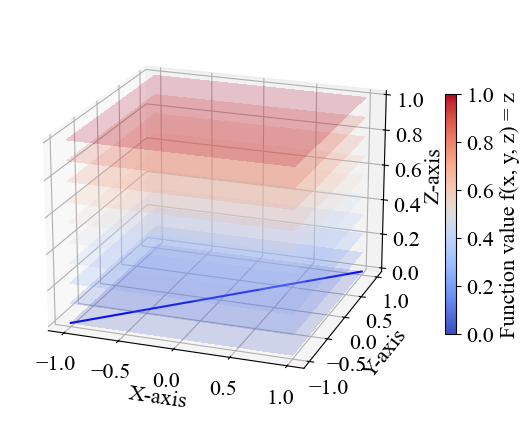}
    \caption*{(a) Random Embedding}
  \end{minipage}
  \hfill
  \begin{minipage}[t]{0.32\textwidth}
    \centering
    \includegraphics[width=\linewidth]{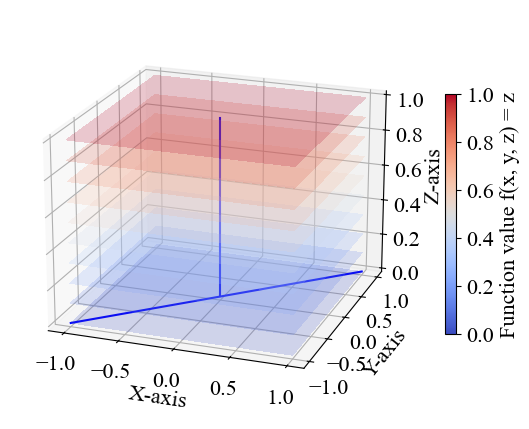}
    \caption*{(b) Cross Embedding Case 1}
  \end{minipage}
  \hfill
  \begin{minipage}[t]{0.32\textwidth}
    \centering
    \includegraphics[width=\linewidth]{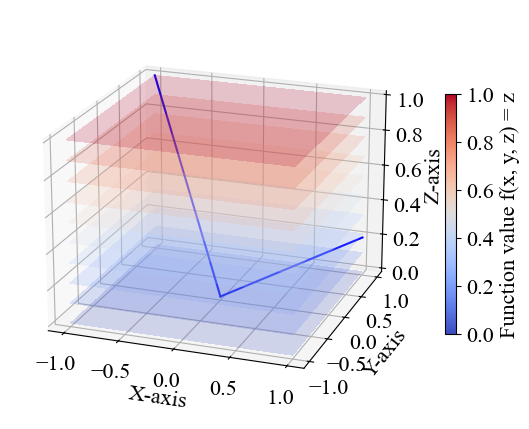}
    \caption*{(c) Cross Embedding Case 2}
  \end{minipage}
  \caption{Random Embeddings in 3D}
    \label{chapt3_fig:3d_cross_random_embedding_ineffective}
\end{figure}

\ac{REMBO} leverages a random projection matrix $A$, which maps points from the high-dimensional space to the low-dimensional space where the optimization occurs. While this approach significantly reduces the dimensionality of the search space, the random nature of the projection matrix $A$ often results in inefficiencies. Specifically, the randomness in the projection can lead to suboptimal exploration of the solution space, limiting the effectiveness of the optimization process.

To mitigate this issue, we propose an enhanced algorithm that introduces an additional embedding step by orthogonalizing the random projection matrix $A$ through Gram–Schmidt process. This orthogonal transform aims to improve both the efficiency and effectiveness of the search by ensuring that the projection matrix explores the subspace in a more structured manner. The proposed algorithm, referred to as cross Random Embedding Bayesian Optimization (\ac{cREMBO}), retains the core benefits of \ac{REMBO} while addressing its limitations through controlled embedding transformations.

\begin{algorithm}[ht]
\caption{\ac{cREMBO}}
\label{alg:cREMBO}
\algcomment{$^{\$}$The acquisition metric is \textit{f}(surrogate model predictions of function value and uncertainty), where \textit{f} is the acquisition function. $^{\&}$orthogonal(A) is based on QR decomposition and Gram–Schmidt process.}
\begin{algorithmic}[1]
    \small
    \State \textbf{Input:} Initial dataset $D_0 = \{(x_i, f(x_i))\}_{i=1}^n$, embedding dimension $d_e$, search region $S$, random projection matrix $A \in \mathbb{R}^{d\times d_e} (d \ge d_e)$, maximum iterations $T$
    \State \textbf{initialization:} 
    \State $\mathcal{D}_0 = D_0$
    \State $t = 0$
    \State Find orthogonal of random matrix $A$: $orthogonal A$
    \While{$t < T$}
        \State Finding $x_{t+1}$ to make $\underset{x_{t+1} \in S}{\mathrm{arg\,max}}$(metric$^{\$}$($x_{t+1}$)) 
        
        \State \# Random Embedding
        \State Measurement of objective function $y_{t+1}=F(A \cdot y_{t+1})$
        \State Updating surrogate model on augmented dataset: $\mathcal{D}_t \cup (x_{t+1}, y_{t+1})$ 
        \State \# Cross Embedding
        \State Measurement of objective function: $y_{t+2} \gets \text{orthogonal}^{\&}(A) \cdot y$
        \State Updating surrogate model on augmented dataset:  $\mathcal{D}_{t+2} \gets \mathcal{D}_t \cup \{(y_{t+2}, \text{feedback\_2})\}$
    \EndWhile

    \State \Return $optimum$
\end{algorithmic}
\end{algorithm}

Because of the randomness of embedding in \ac{REMBO}, different embeddings influence the efficiency and effectiveness of \ac{REMBO}. And random embedding has embedding distortion issue, i.e., the distances between two points before and after embedding are different. Figure \ref{chapt3_fig:3d_cross_random_embedding_ineffective} (a) shows a example of random embedding, because of the randomness in Random Embedding of \ac{REMBO},the \ac{REMBO} would assume the effective dimension is Z-axis based on the sampling points on blue line for this contour function and pay more attention on X dimension, which is a waste of computational cost (inefficient) and the random embedding (the blue line) may miss the global optimum searching along the embedding (effectiveness). If the sampling bias from random sampling can be compensated and the coverage of sampling can be improved by adding an embedding that is 'orthogonal' to the random embedding, the \ac{REMBO} can be improved. Figure \ref{show 2d cross embedding} and Figure \ref{chapt3_fig:3d_cross_random_embedding_ineffective} (b), (c) illustrates how 2D and 3D examples of \ac{cREMBO} operate. In these examples, the added orthogonal embedding enables complementary and more efficient exploration by expanding the search into multiple subspaces. Each projection matrix defines a distinct subspace basis, potentially introducing a distortion or scaling along those basis directions, allowing the optimizer to discover optima more effectively across varying directions. This approach is particularly advantageous for nonstationary functions, where the objective exhibits different variation patterns or multiple optima in different regions of the search space.

As discussed in Section \ref{sec:REMBO}, \ac{REMBO} uses a bounded region in low dimension before projecting to high dimensions. Without a bounded region, the random embedding could project points far outside the feasible region in the high-dimensional space, leading to more necessary warping of infeasible solution outside the search space to boundary that worsens existing over-exploration on bounds issue of \ac{REMBO} \cite{wang2016bayesian}. Note that the \ac{REMBO} algorithm does not guarantee that the global optimum of the high-dimensional objective function lies within the projected region corresponding to the bounded box in the low-dimensional space \cite{letham2020re}. Instead, it assumes that the high-dimensional function can be effectively represented in a lower-dimensional subspace, with a probability that depends on the choice of embedding and the dimensionality of the subspace. \cite{wang2013bayesian} 

Though we can also use a scaling of embedding method to improve effectiveness of embedding in \ac{REMBO}, this also makes the boundaries over-exploration issue of \ac{REMBO} worse \cite{wang2016bayesian} (more sampled points outside the box projected to the boundaries), and it does not help jump out of local optimum by correcting the bias from random embedding, which is the drawback of \ac{REMBO}. 

The Algorithm \ref{alg:cREMBO} shows details about \ac{cREMBO}. The difference with \ac{REMBO} is that it adds an orthogonal embedding in addition to the random embedding. It is a general and simple method that to improve \ac{REMBO}. To get the orthogonal embedding, the random embedding in high dimensions ($A\cdot y_{t+1}$, where $y_{t+1}$ is low dimension sample) is rotated to the orthogonal angle and traced back to low dimensions as a new sample point ($y_{t+2}$).

Let $\{A_1, A_2, \ldots, A_K\}$ denote $K$ random embedding matrices, where each $A_k \in \mathbb{R}^{D \times d}$. For each embedding $k \in \{1, \ldots, K\}$ and low-dimensional input $x \in \mathbb{R}^d$, we define the projected point in the original high-dimensional space as:
\[
\tilde{x}_k = A_k x.
\]
Instead of treating each $\tilde{x}_k$ independently, we define a single GP model over the augmented input space $(x, z) \in \mathbb{R}^d \times \mathcal{Z}$, where $z \in \{1, \ldots, K\}$ is a discrete embedding index. The GP surrogate then models:
\[
f(x, z) := f(A_z x).
\]

\subsection{Kernel Construction}

In our cREMBO framework, we model the joint input space comprising the embedding index $z \in \{1, \ldots, K\}$ and the embedded optimization variable $x \in \mathbb{R}^d$ using a product kernel of the form $k((x_i, z_i), (x_j, z_j)) = k_x(x_i, x_j) \cdot k_z(z_i, z_j)$. For the continuous input space, $k_x$ is chosen as an ARD kernel (e.g., Matern or squared exponential), where a separate lengthscale is learned for each input dimension to capture anisotropic sensitivity. 

For the discrete embedding index $z$, we adopt a smooth exponential kernel defined as $k_z(z_i, z_j) = \exp(-\lambda^2 (z_i - z_j)^2)$, which corresponds to a Gaussian process over the index space. This avoids artificially injecting discrete distances or one-hot encodings that introduce magnitude discontinuities and interfere with automatic relevance detection. By placing the embedding index and functional input in a joint kernel space with shared hyperparameter tuning, the model can learn which embeddings are relevant in which regions of the domain. This formulation enables adaptive selection among multiple subspaces and mitigates the limitations of using a single projection in standard REMBO.

The GP kernel $K((x_i, z_i), (x_j, z_j))$ over the augmented input is factorized as a product kernel:
\[
K((x_i, z_i), (x_j, z_j)) = K_x(x_i, x_j) \cdot K_z(z_i, z_j),
\]
where:
\begin{itemize}
    \item $K_x$ is a standard kernel over $\mathbb{R}^d$, such as the Matern-5/2 or squared exponential kernel.
    \item $K_z$ is a kernel defined over the discrete embedding index space $\mathcal{Z}$. Several constructions are possible:
    \begin{enumerate}
        \item \textbf{Delta Kernel (No Sharing):} $K_z(z_i, z_j) = \delta_{z_i = z_j}$.
        \item \textbf{Learned Embedding Kernel:} Each index $z$ is mapped to a latent vector $\phi(z) \in \mathbb{R}^r$ and $K_z(z_i, z_j) = k(\phi(z_i), \phi(z_j))$.
        \item \textbf{Tanimoto or Categorical Kernel:} For simple categorical treatment with task similarity.
    \end{enumerate}
\end{itemize}

This setup enables the model to learn correlations across embeddings. It can emphasize embeddings that better represent the function structure while down-weighting others.

\subsection{Acquisition Optimization}

To optimize the acquisition function $\alpha(x, z)$ over both $x$ and $z$, we must handle the mixed-variable nature of the augmented domain. Several approaches are possible:

\begin{itemize}
    \item \textbf{Exhaustive enumeration over $z$:} Evaluate $\max_{x \in \mathcal{X}} \alpha(x, z)$ for each $z$, and select the best.
    \item \textbf{Mixed-integer Evolution Strategies (MIES):} Jointly optimize over continuous $x$ and discrete $z$ using evolutionary algorithms.
    \item \textbf{One-hot embedding with continuous relaxation:} Treat $z$ as a continuous latent vector and use Gumbel-softmax or similar techniques.
\end{itemize}

\subsection{Theoretical Implications}

By jointly modeling all embeddings, the GP surrogate benefits from increased sample efficiency, as information from all queries contributes to a single model. This contrasts with traditional REMBO, where samples from different embeddings are disjoint in model space.

The method also provides a mechanism for learning a posterior over the relevance of embeddings. If some $A_k$ capture the true function subspace better, the GP will assign higher predictive confidence in those regions via learned correlations in $K_z$.

\subsection{Advantages and Limitations}

\paragraph{Advantages}
\begin{itemize}
    \item \textbf{Sample Efficiency:} All evaluated points contribute to a unified model, improving learning rates.
    \item \textbf{Correlation Exploitation:} The GP kernel can learn to correlate similar embeddings, effectively performing embedding selection.
    \item \textbf{Scalability:} The model scales to many embeddings without needing $K$ independent GPs.
    \item \textbf{Robustness:} Averaging over multiple embeddings mitigates the failure of any single projection.
\end{itemize}

\paragraph{Limitations}
\begin{itemize}
    \item \textbf{Kernel Design for $z$:} Modeling the discrete index $z$ is nontrivial and sensitive to kernel choice.
    \item \textbf{Optimization Complexity:} The acquisition function requires specialized mixed-variable optimization techniques.
    \item \textbf{Embedding Interaction Complexity:} If the embeddings are too dissimilar, the model may suffer from conflicting data.
\end{itemize}

\section{Experiments}
\subsection{styblinskiTang Function}
We evaluate on the 8-dimensional Styblinski–Tang function, a highly multimodal and non-convex benchmark defined as a sum of fourth-order polynomial terms. Although separable, its rugged landscape with numerous local minima and varying curvature across dimensions introduces non-stationarity when embedded into high-dimensional spaces. This makes it well-suited to test the robustness of embedding-based optimization methods, particularly in scenarios where a single global projection is insufficient. This function also reflects real-world optimization challenges such as hyperparameter tuning for machine learning models, where the loss landscape is rugged with many deceptive minima, and analog circuit design, where different subsets of parameters are influential in different regions of the design space. Therefore, the Styblinski–Tang function serves as a strong empirical benchmark to test an optimizer’s ability to adapt to sparsity, non-stationarity, and multimodality in high-dimensional Bayesian optimization.

\section{Method Description}

\begin{figure}[t]
  \centering
  \begin{minipage}[t]{0.49\textwidth}
    \centering
    \includegraphics[width=\linewidth]{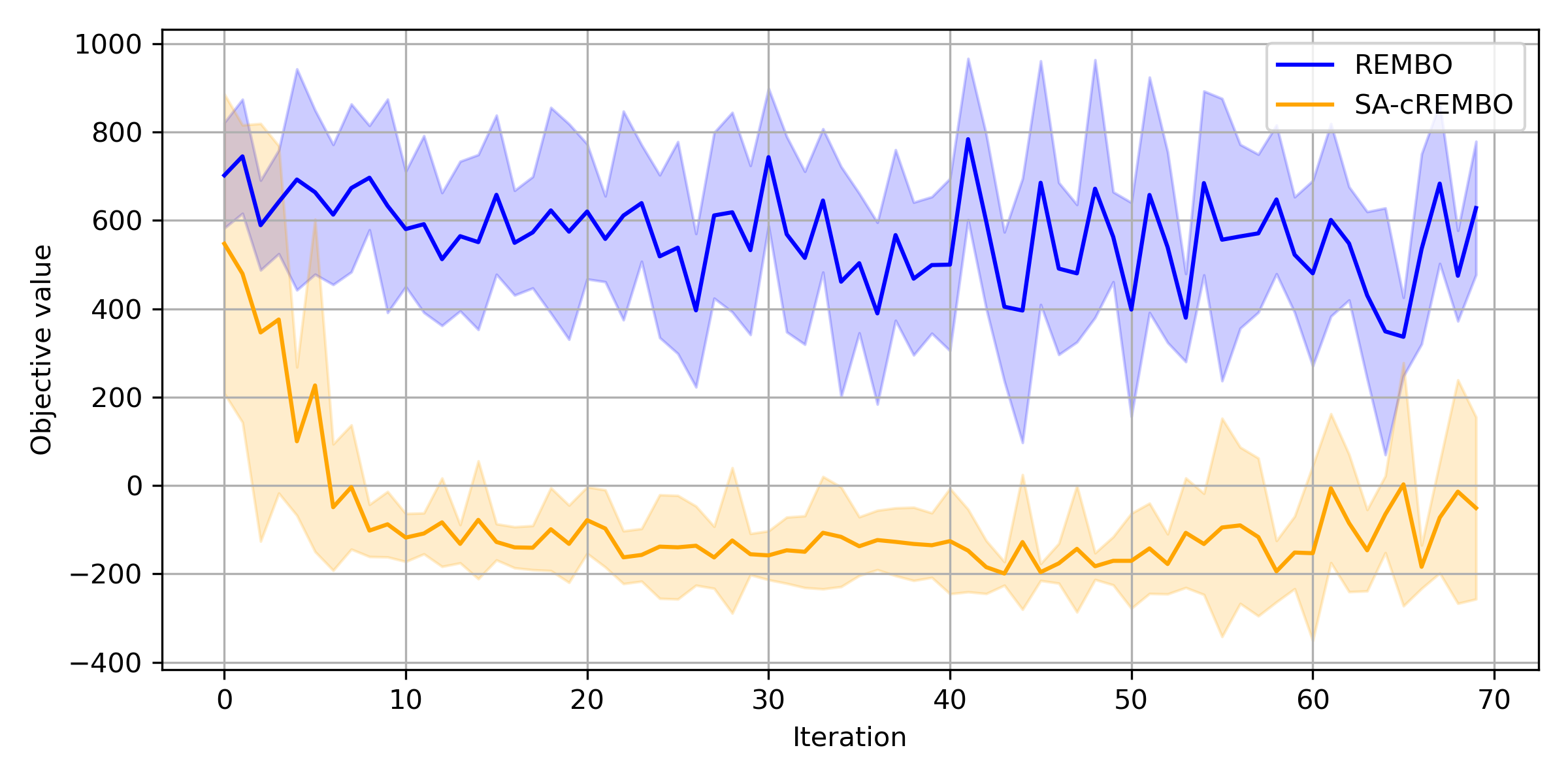}
    \caption*{(a) Styblinski–Tang Function (d=8 D =21)}
  \end{minipage}
  \hfill
  \begin{minipage}[t]{0.49\textwidth}
    \centering
    \includegraphics[width=\linewidth]{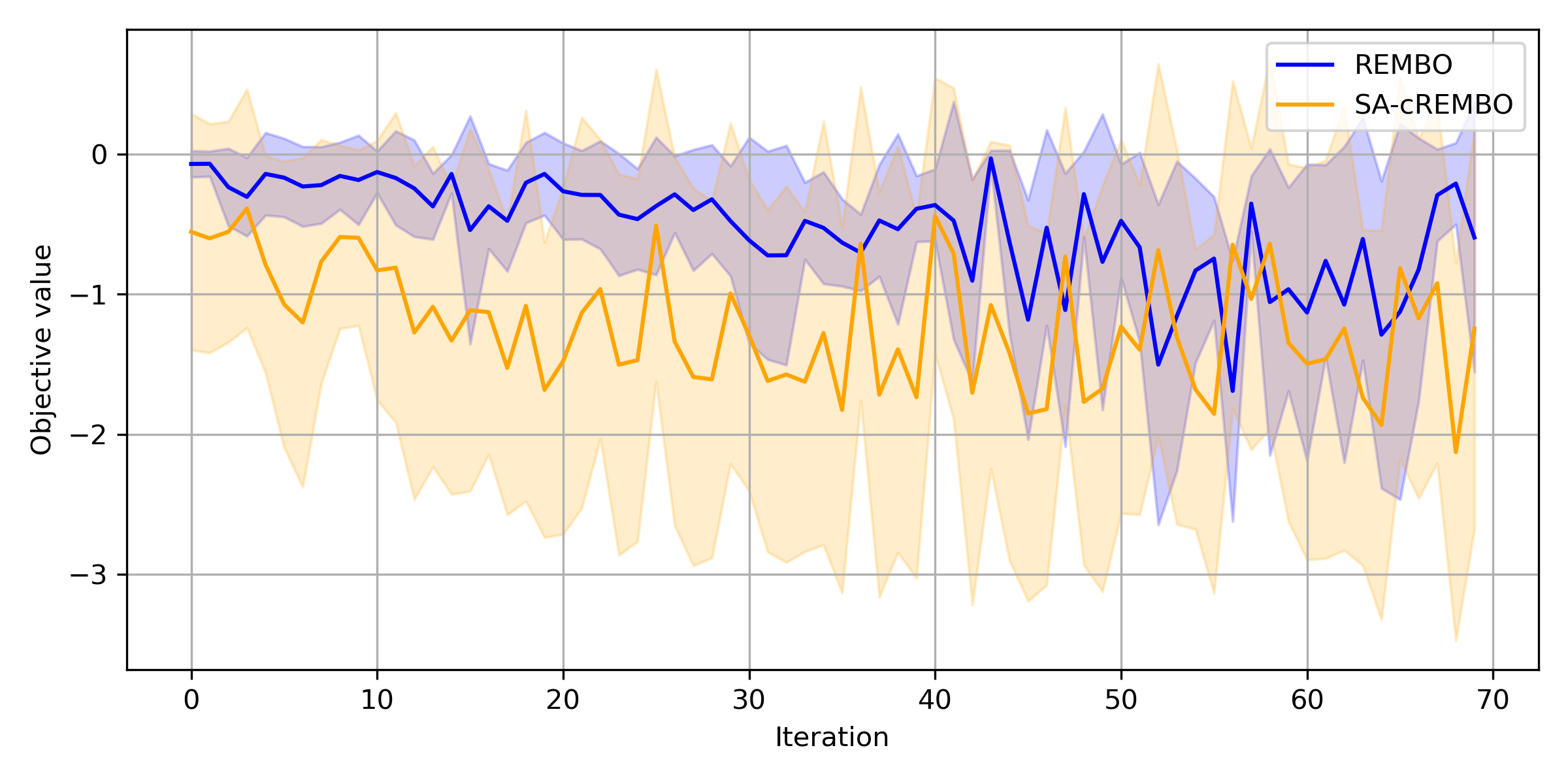}
    \caption*{(b) Harzmann6 Function (d=6 D =21)}
  \end{minipage}
  \caption{Results comparison of different optimizers on Styblinski–Tang and Harzmann6 functions.}
  \label{fig:optimizer_comparisons}
\end{figure}

\begin{figure}[htbp]
\centering
\includegraphics[width=1\textwidth]{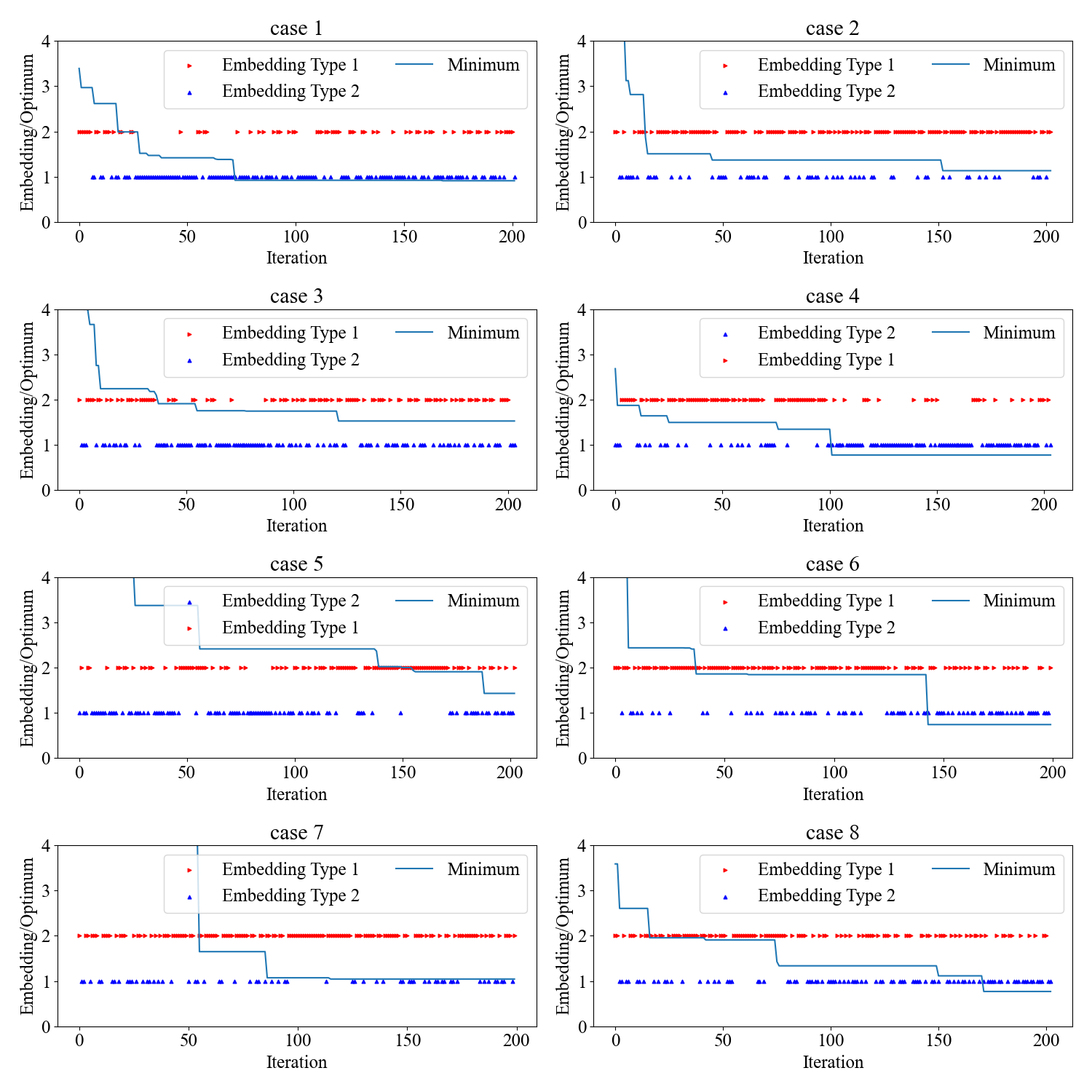}
\caption{The Automatic Choice of Embedding in \ac{cREMBO}.\label{fig:auto choose of embedding}}
\end{figure}


\begin{figure}[htbp]
\centering
\includegraphics[width=1\textwidth]{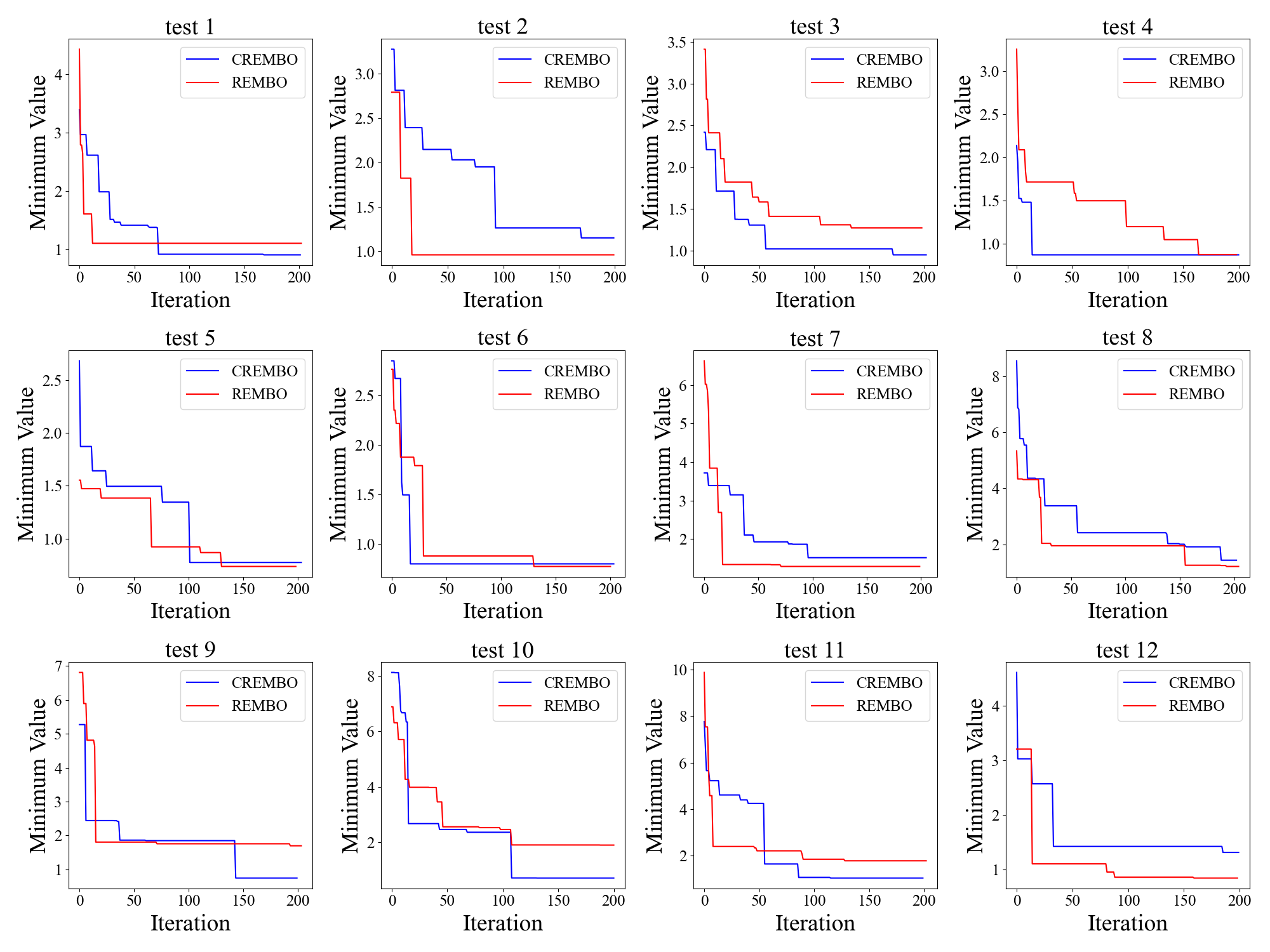}
\caption{The Optimization Results Comparisons of \ac{REMBO} and \ac{cREMBO}.\label{chapt3_crembo_vs_rembo}}
\end{figure}

\begin{figure}[t]
  \centering
  \begin{minipage}[t]{0.4\textwidth}
    \centering
    \includegraphics[width=\linewidth]{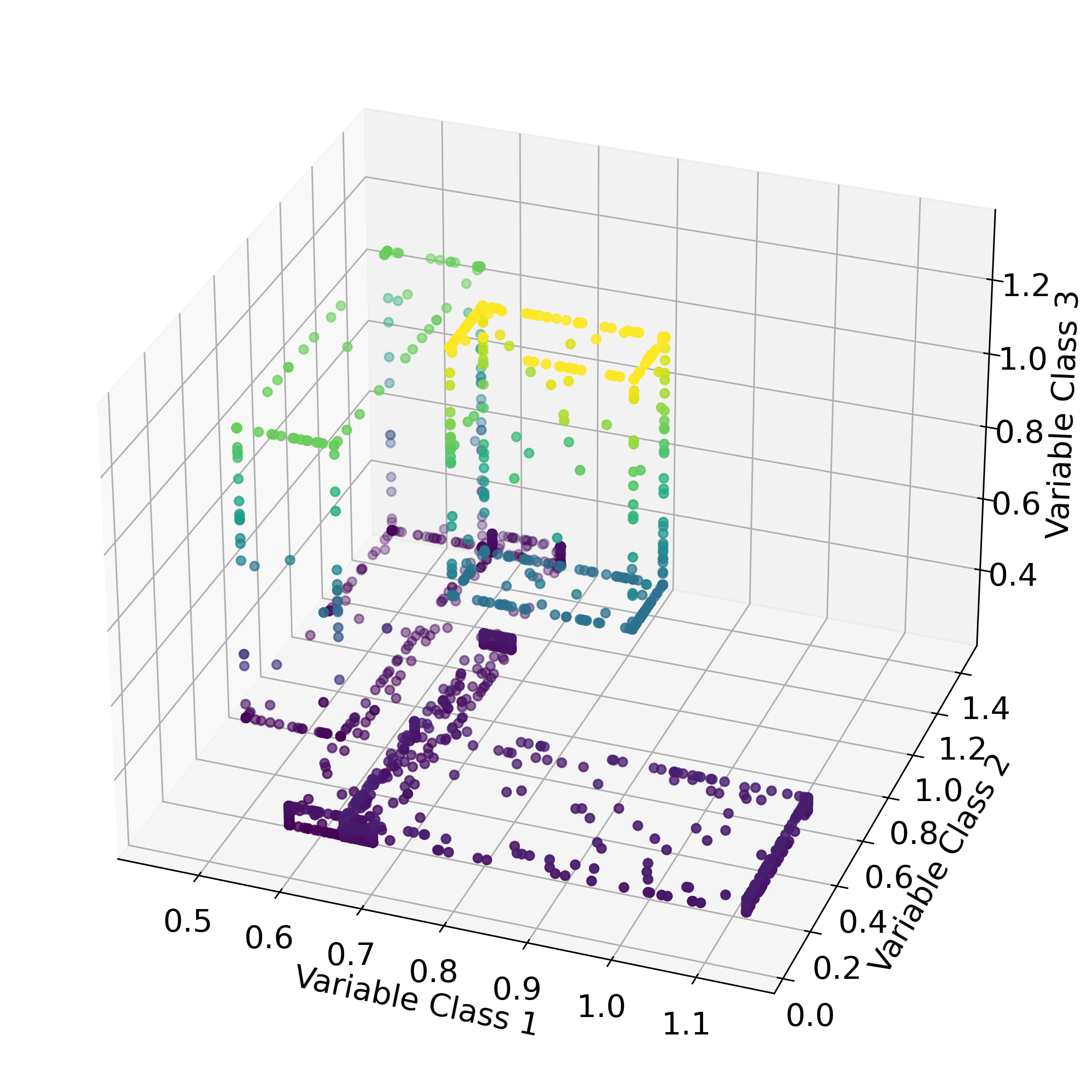}
    \caption*{(a) \ac{REMBO}}
  \end{minipage}
  \hfill
  \begin{minipage}[t]{0.4\textwidth}
    \centering
    \includegraphics[width=\linewidth]{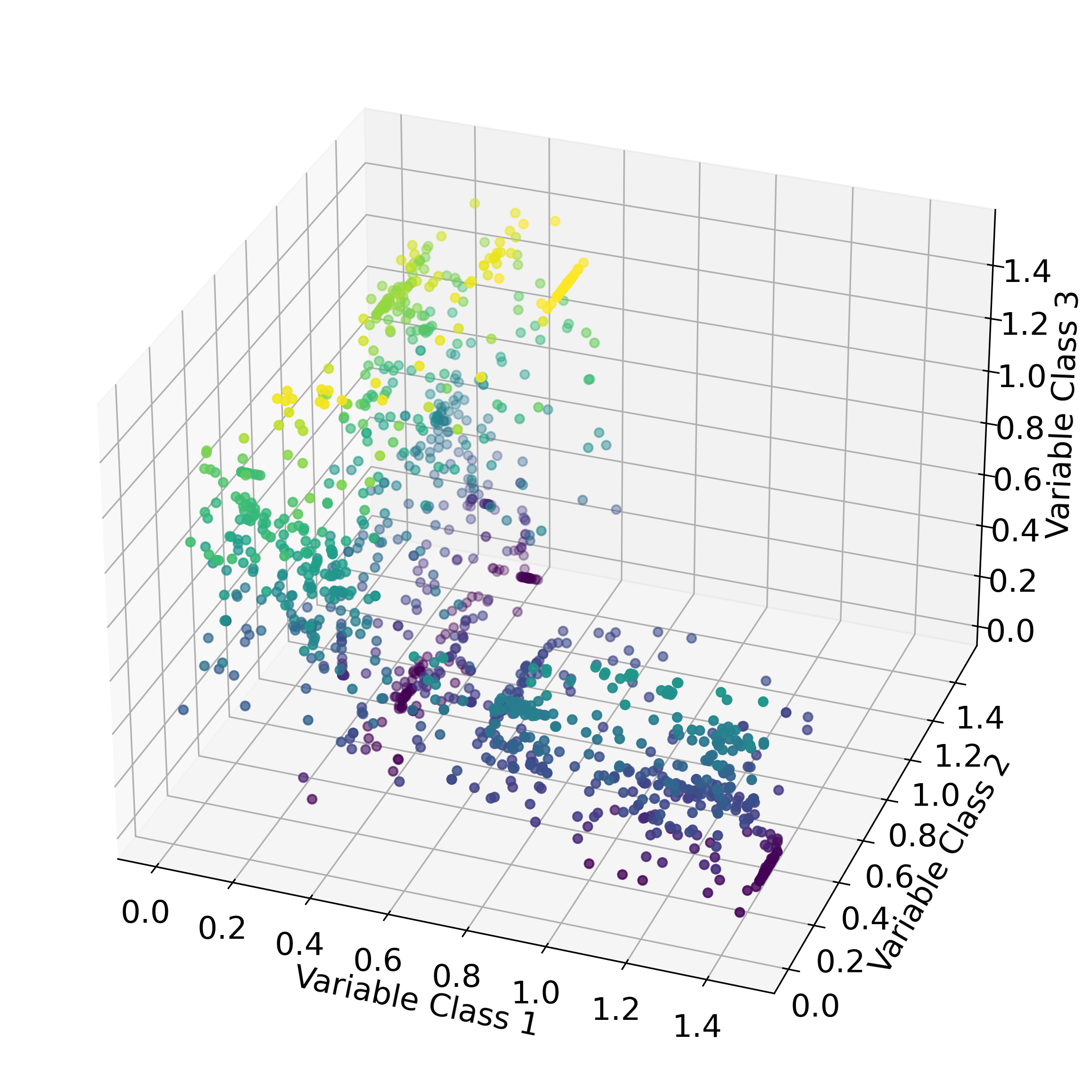}
    \caption*{(b) \ac{cREMBO}}
  \end{minipage}
  
  \caption{The Comparison of Projected Sample Distributions in \ac{REMBO} and \ac{cREMBO}.}
  \label{fig:REMBO_vs_cREMBO}
\end{figure}

\section{Conclusion}

This unified surrogate modeling framework extends REMBO by enabling multiple embeddings to be incorporated into a single GP using an additional index dimension. The proposed method combines the diversity benefits of ensemble REMBO with the efficiency of joint modeling, offering a scalable and flexible alternative for high-dimensional Bayesian Optimization. Its effectiveness depends on proper kernel design, especially over the embedding index, and the availability of computational resources to train expressive GP models over mixed domains.

\section{Supplementary Material}

\textbf{Proof. of Theorem 1.} 
Since \( f \) has an effective dimensionality \( d_e \), there exists a subspace \( T \subset \mathbb{R}^D \), such that \( \text{rank}(T) = d_e \). Furthermore, any \( \bm{x} \in \mathbb{R}^D \) can be decomposed as \( \bm{x} = \bm{x}_{\top} + \bm{x}_{\perp} \), where \( \bm{x}_{\top} \in T \) and \( \bm{x}_{\perp} \in T^\perp \) (the orthogonal complement of \( T \)). Therefore, we have
\begin{equation}
f(\bm{x}) = f(\bm{x}_{\top} + \bm{x}_{\perp}) = f(\bm{x}_{\top}),
\label{eq:f_decomposition}
\end{equation}
which implies that the function \( f \) only depends on \( \bm{x}_{\top} \), the component of \( \bm{x} \) in the subspace \( T \). Thus, it suffices to show that for all \( \bm{x}_{\top} \in T \), there exists a \( \bm{y} \in \mathbb{R}^d \) such that
\begin{equation}
f(\bm{x}_{\top}) = f(\bm{A}\bm{y}).
\label{eq:f_xy}
\end{equation}

Let \( \Phi \in \mathbb{R}^{D \times d_e} \) be a matrix whose columns form an orthonormal basis for \( T \). For each \( \bm{x}_{\top} \in T \), there exists a \( \bm{c} \in \mathbb{R}^{d_e} \) such that
\begin{equation}
\bm{x}_{\top} = \Phi \bm{c}.
\label{eq:x_top}
\end{equation}

Assuming that \( \Phi^\top \bm{A} \) has rank \( d_e \), we know there exists a \( \bm{y} \in \mathbb{R}^d \) such that
\begin{equation}
(\Phi^\top \bm{A}) \bm{y} = \bm{c}.
\label{eq:Phi_A_y_c}
\end{equation}

The orthogonal projection of \( \bm{A}\bm{y} \) onto \( T \) is given by
\begin{equation}
\Phi\Phi^\top \bm{A} \bm{y} = \Phi \bm{c} = \bm{x}_{\top}.
\label{eq:projection}
\end{equation}
Thus, we have \( \bm{A} \bm{y} = \bm{x}_{\top} + \bm{x}_{\perp} \) for some \( \bm{x}_{\perp} \in T^\perp \), since \( \bm{x}_{\top} \) is the projection of \( \bm{A} \bm{y} \) onto \( T \). Consequently,
\begin{equation}
f(\bm{A} \bm{y}) = f(\bm{x}_{\top} + \bm{x}_{\perp}) = f(\bm{x}_{\top}),
\label{eq:f_Ay_x_top}
\end{equation}
which completes the equivalence.

Now, it remains to show that with probability 1, the matrix \( \Phi^\top \bm{A} \) has rank \( d_e \). Let \( \bm{A}_e \in \mathbb{R}^{D \times d_e} \) be a submatrix of \( \bm{A} \), consisting of any \( d_e \) columns of \( \bm{A} \), which are independent and identically distributed (i.i.d.) samples drawn from \( \mathcal{N}(0, \bm{I}) \). Then, the vectors \( \Phi^\top \bm{a}_i \) are i.i.d. samples from the distribution
\begin{equation}
\mathcal{N}\left( \bm{0}, \Phi^\top \Phi \right) = \mathcal{N}\left( \bm{0}_{d_e}, \bm{I}_{d_e \times d_e} \right),
\label{eq:normal_distribution}
\end{equation}
and hence \( \Phi^\top \bm{A}_e \), when considered as an element of \( \mathbb{R}^{d_e^2} \), is a sample from
\begin{equation}
\mathcal{N}\left( \bm{0}_{d_e^2}, \bm{I}_{d_e^2 \times d_e^2} \right).
\label{eq:Phi_A_e_distribution}
\end{equation}

Since the set of singular matrices in \( \mathbb{R}^{d_e^2} \) has Lebesgue measure zero (as it is the zero set of a polynomial, such as the determinant function), and polynomial functions are Lebesgue measurable, the matrix \( \Phi^\top \bm{A}_e \) is almost surely non-singular. Therefore, it has rank \( d_e \), and the same is true for \( \Phi^\top \bm{A} \), whose columns include the columns of \( \Phi^\top \bm{A}_e \).

\textbf{Proof of Theorem 2.} 
Since \( \bm{X} \) is a box constraint, projecting \( \bm{x}^* \) onto \( \bm{T} \) ensures that \( \bm{x}^* \in \bm{T} \cap \bm{X} \). Moreover, as \( \bm{x}^* = \bm{x}^*_\top + \bm{x}_\perp \) for some \( \bm{x}_\perp \in \bm{T}^\perp \), we have \( f(\bm{x}^*) = f(\bm{x}^*_\top) \). Therefore, \( \bm{x}^*_\top \) is an optimizer. Following the argument from Proposition 1, it holds with probability 1 that for all \( \bm{x} \in \bm{T} \), there exists a \( \bm{y} \in \mathbb{R}^d \) such that \( \bm{A}\bm{y} = \bm{x} + \bm{x}_\perp \) where \( \bm{x}_\perp \in \bm{T}^\perp \). Let \( \Phi \) be the matrix whose columns form a standard basis for \( \bm{T} \). Without loss of generality, assume that
\begin{equation}
\Phi = \begin{bmatrix} 
\bm{I}_{d_e} & \bm{0} 
\end{bmatrix}.
\label{eq:Phi_basis}
\end{equation}
Then, as shown in Proposition \ref{eq:projection}, there exists a \( \bm{y}^* \in \mathbb{R}^d \) such that
\begin{equation}
\Phi\Phi^\top \bm{A}\bm{y}^* = \bm{x}^*_\top.
\label{eq:Phi_A_y_star}
\end{equation}
For each column of \( \bm{A} \), we have
\begin{equation}
\Phi\Phi^\top \bm{a}_i \sim \mathcal{N}\left( 
\bm{0}, \begin{bmatrix} 
\bm{I}_{d_e} & \bm{0} \\
\bm{0} & \bm{0} 
\end{bmatrix}
\right).
\label{eq:Gaussian_distribution}
\end{equation}
Therefore, \( \Phi\Phi^\top \bm{A}\bm{y}^* = \bm{x}^*_\top \) is equivalent to
\begin{equation}
\bm{B}\bm{y}^* = \bar{\bm{x}}^*_\top
\label{eq:B_y_star}
\end{equation}
where \( \bm{B} \in \mathbb{R}^{d_e \times d_e} \) is a random matrix with independent standard Gaussian entries, and \( \bar{\bm{x}}^*_\top \) contains the first \( d_e \) entries of \( \bm{x}^*_\top \) (the remaining entries are zero). By Theorem 3.4 from \cite{sankar2006smoothed}, we have
\begin{equation}
\Pr\left( \| \bm{B}^{-1} \|_2 \geq \sqrt{d_e} \right) \leq \epsilon.
\label{eq:prob_bound_B_inv}
\end{equation}
Thus, with probability at least \( 1-\epsilon \),
\begin{equation}
\| \bm{y}^* \| \leq \| \bm{B}^{-1} \|_2 \| \bar{\bm{x}}^*_\top \|_2 = \| \bm{B}^{-1} \|_2 \| \bm{x}^*_\top \|_2 \leq \sqrt{d_e} \epsilon \| \bm{x}^*_\top \|_2.
\label{eq:bound_y_star}
\end{equation}

\bibliographystyle{IEEEtran}

\bibliography{neurips_2023}

\end{document}

%% file: 1_packages.tex
\usepackage[Rejne]{fncychap}

\usepackage[
  pdfauthor={Yuejiang Wen},
  pdftitle={High Dimensional Optimization for Electronic Designs},
  pdfcreator={pdftex},
  pdfsubject={NC State ETD Dissertation},
  pdfkeywords={High dimension, random embedding, Bayesian optimization, cross embedding, analog, signal integrity, acquisition function, hybrid kernel},
  colorlinks=false,
  linkcolor=black,
  citecolor=black,
  filecolor=black,
  urlcolor=black,
]{hyperref}

\usepackage{amsmath,amssymb,amsfonts,amsthm,algorithm,array}
\usepackage{subcaption}

\usepackage{textcomp,stfloats,url,verbatim,graphicx,svg}
\usepackage{tabu,makecell,multirow,bm,cuted,lipsum} 
\usepackage{textcomp,xcolor,algpseudocode}
\usepackage{boldline,caption,comment,color,diagbox}
\usepackage{fancyhdr,floatflt,float}
\usepackage{indentfirst,latexsym,lineno,mathtools}
\usepackage{picinpar,placeins,psfrag,physics,romannum,stfloats,soul}
\usepackage{subcaption}
\usepackage{graphicx}

\usepackage{xcolor}

\usepackage{wrapfig}

\usepackage{microtype}
\usepackage[T1]{fontenc}
\usepackage{amssymb,amsfonts}
\usepackage{array}
\usepackage{boxedminipage}
\usepackage{boldline}

\usepackage{comment}
\usepackage{color}
\usepackage{diagbox}
\usepackage{fancyhdr}
\usepackage{floatflt}
\usepackage{float}
\usepackage{graphics}
\usepackage{graphicx} 
\usepackage{hyperref}
\hypersetup{colorlinks=false,linkcolor=black,filecolor=black,urlcolor=black,citecolor=black}
\usepackage{indentfirst}
\usepackage{latexsym}
\usepackage{lineno}
\usepackage{mathtools}
\usepackage{makecell}
\usepackage{multirow}
\usepackage{picinpar}
\usepackage{placeins}
\usepackage{psfrag}
\usepackage{physics}
\usepackage{romannum}
\usepackage{stfloats}
\usepackage{soul}
\usepackage{tabu}
\usepackage{textcomp}
\usepackage{theorem}
\usepackage{threeparttable}

\usepackage{url}
\usepackage{xcolor}
\usepackage{subcaption}

\usepackage{booktabs}  

\usepackage{textcomp}  
\usepackage{xcolor}
\usepackage{lipsum}    
\usepackage{longtable}

\makeatletter
\AfterEndEnvironment{algorithm}{\let\@algcomment\relax}
\AtEndEnvironment{algorithm}{\kern2pt\hrule\relax\vskip3pt\@algcomment}
\let\@algcomment\relax
\newcommand\algcomment[1]{\def\@algcomment{\footnotesize#1}}

\renewcommand\fs@ruled{\def\@fs@cfont{\bfseries}\let\@fs@capt\floatc@ruled
  \def\@fs@pre{\hrule height.8pt depth0pt \kern2pt}%
  \def\@fs@post{}%
  \def\@fs@mid{\kern2pt\hrule\kern2pt}%
  \let\@fs@iftopcapt\iftrue}
\makeatother

\DeclareUnicodeCharacter{2212}{-}

%% file: 2_acronyms.tex
\usepackage{acronym}
\newacro{45RFSOI}  [45RFSOI] {GlobalFoundries 45\,nm RFSOI CMOS}
\newacro{90}       [$90^\circ$] {$90^\circ$}
\newacro{180}      [$180^\circ$] {$180^\circ$}
\newacro{360}      [$360^\circ$] {$360^\circ$}
\newacro{4g}       [4\sc{g}] {fourth-generation}
\newacro{5g}       [5\sc{g}] {fifth-generation}
\newacro{6g}       [6\sc{g}] {sixth-generation}
\newacro{ADI}      [ADI] {Analog Devices Inc.}
\newacro{ADC}      [ADC] {analog-to-digital converter}
\newacro{AM}       [AM] {amplitude modulation}
\newacro{API}      [API] {application programming interface}
\newacro{BB}       [BB] {baseband}
\newacro{BBVGA}    [BBVGA] {baseband variable-gain amplifier}
\newacro{BFIC}     [BFIC] {beamforming IC}
\newacro{BIST}     [BIST] {built-in self-test}
\newacro{BS}       [BS] {basestation}
\newacro{BPSK}     [BPSK] {binary phase-shift keying}
\newacro{BO}       [BO] {Bayesian optimization}
\newacro{BW}       [BW] {bandwidth}
\newacro{CG}       [CG] {conversion gain}
\newacro{CDMA}     [CDMA] {code-division multiple access}
\newacro{COM}      [COM] {component object model}
\newacro{CoMET}    [CoMET] {code-modulated embedded test}
\newacro{COTS}     [COTS] {commercial off-the-shelf}
\newacro{CMI}      [CMI] {code-modulated interferometry}
\newacro{CMOS}     [CMOS] {complementary metal–oxide–semiconductor}
\newacro{CU}       [CU] {central unit}
\newacro{DAC}      [DAC] {digital-to-analog converter}
\newacro{DFE}      [DFE] {digital front-end}
\newacro{DR}       [DR] {dynamic range}
\newacro{DU}       [DU] {distributed unit}
\newacro{ECE}      [ECE] {Electrical and Computer Engineering}
\newacro{EIRP}     [EIRP] {effective isotropic radiated power}
\newacro{EMI}      [EMI] {electromagnetic interference}
\newacro{EVM}      [EVM] {error vector magnitude}
\newacro{fpga}     [FPGA] {field programmable gate array}
\newacro{FR2}      [\sc{Fr2}]  {frequency-range two}
\newacro{FR3}      [\sc{Fr3}]  {frequency-range three}
\newacro{fspl}     [FSPL] {free-space path loss}
\newacro{GF}       [GF] {GlobalFoundries}
\newacro{Gm}       [G\textsubscript{m}] {transconductance}
\newacro{GP}       [GP] {Gaussian process}
\newacro{H}        [H] {horizontal}
\newacro{I}        [$I$] {in-phase}
\newacro{IB}       [IB] {in-band}
\newacro{iBW}      [iBW] {instantaneous bandwidth}
\newacro{IC}       [IC] {integrated circuit}
\newacro{IF}       [IF] {intermediate frequency}
\newacro{IIP3}     [iIP\textsubscript{3}] {input-referred third-order intercept point}
\newacro{IM3}      [IM\textsubscript{3}]{third-order intermodulation product}
\newacro{iP1dB}    [iP\textsubscript{1dB}] {input-referred 1 dB compression point}
\newacro{iPLSB}    [iP\textsubscript{LSB}] {input-referred phase compression point by one LSB}
\newacro{IR}       [IR] {image-reject}
\newacro{LNA}      [LNA] {low-noise amplifier}
\newacro{LNTA}     [LNTA] {low-noise transconductance amplifier}
\newacro{LO}       [LO] {local oscillator}
\newacro{LOS}      [L\sc{o}S] {line-of-sight}
\newacro{LSB}      [LSB] {least-significant bit}
\newacro{LUT}      [LUT] {look-up table}
\newacro{MIMO}     [MIMO] {multiple-input multiple-output}
\newacro{ML}       [ML] {machine learning}
\newacro{mmwave}   [mmWave] {millimeter-wave}
\newacro{NCSU}     [NC State] {North Carolina State University}
\newacro{NI}       [NI] {National Instruments}
\newacro{NLOS}      [NL\sc{o}S] {non line-of-sight}
\newacro{NF}       [NF] {noise figure}
\newacro{NR}       [NR] {new radio}
\newacro{NSF}      [NSF] {National Science Foundation}
\newacro{OC}       [O.C.] {open circuit}
\newacro{OCU}      [O-CU] {open centralized unit}
\newacro{ODU}      [O-DU] {open distributed unit}
\newacro{OEM}      [OEM] {original equipment manufacturer}
\newacro{OFDM}     [OFDM] {orthogonal frequency-division multiplexing}
\newacro{OOB}      [OOB] {out-of-band}
\newacro{opamp}    [op-amp] {operational amplifier}
\newacro{ORAN}     [O-RAN] {open radio-access network}
\newacro{ORU}      [O-RU] {open radio unit}
\newacro{OTA}      [OTA] {over-the-air}
\newacro{pcb}      [PCB] {printed-circuit board}
\newacro{oP1dB}    [oP\textsubscript{1dB}] {output-referred 1 dB compression point}
\newacro{oPLSB}    [oP\textsubscript{LSB}] {output-referred phase compression point by one LSB}
\newacro{PA}       [PA] {power amplifier}
\newacro{PAE}      [PAE] {power-added efficiency}
\newacro{PCB}      [PCB] {printed circuit board}
\newacro{PLL}      [PLL] {phase-locked loop}
\newacro{PLSB}     [P\textsubscript{LSB}] {phase compression point by one LSB}
\newacro{PM}       [PM] {phase modulation}
\newacro{PNAX}     [PNAX] {Performance Network Analyzer}
\newacro{Psat}     [P\textsubscript{sat}] {saturated output power}
\newacro{Q}        [$Q$] {quality-factor}
\newacro{QAM}      [QAM] {quadrature amplitude modulation}
\newacro{RF}       [RF] {radio-frequency}
\newacro{RFFE}     [RFFE] {radio-frequency front-end}
\newacro{rfsoc}    [RFSoC] {radio-frequency system-on-chip}
\newacro{RMNF}     [RMNF] {Reflection-Mode N-path Filter}
\newacro{RMS}      [RMS] {root-mean squared}
\newacro{RMSE}      [RMSE] {root-mean squared error}
\newacro{RTPS}     [RTPS] {reflection-type phase shifter}
\newacro{rx}       [R\sc{x}] {receiver}
\newacro{S11}      [S\textsubscript{11}] {input reflection coefficient with output terminated}
\newacro{S21}      [S\textsubscript{21}] {transducer gain}
\newacro{S12}      [S\textsubscript{12}] {reverse transducer gain}
\newacro{S22}      [S\textsubscript{22}] {output reflection coefficient}
\newacro{SATCOM}   [SATCOM] {satellite communication}
\newacro{SC}       [S.C.] {short circuit}
\newacro{SDA}      [S\sc{d}A]  {software-defined array}
\newacro{SDR}      [S\sc{d}R]  {software-defined radio}
\newacro{SNR}      [SNR] {signal-to-noise ratio}
\newacro{soc}      [SoC] {system-on-chip}
\newacro{SOI}      [SOI] {silicon-on-insulator}
\newacro{STPS}     [STPS] {switched-type phase shifter}
\newacro{TIA}      [TIA] {transimpedance amplifier}
\newacro{TR}       [T/R] {transmit/receive}
\newacro{TRM}      [TRM] {transmit/receive module}
\newacro{trx}      [TR\sc{x}] {transceiver}
\newacro{tx}       [T\sc{x}] {transmitter}
\newacro{UE}       [UE] {user equipment}
\newacro{USRP}     [USRP] {Universal Software Radio Peripheral}
\newacro{V}        [V] {vertical}
\newacro{VCO}      [VCO] {voltage-controlled oscillator}
\newacro{VGA}      [VGA] {variable-gain amplifier}
\newacro{VNA}      [VNA] {vector network analyzer}
\newacro{REMBO}    [REMBO] {Random EMbedding Bayesian Optimization}
\newacro{IC-REMBO} [IC-REMBO] {inspection-based combo REMBO}
\newacro{cREMBO}   [cREMBO] {cross random embedding Bayesian optimization}
\newacro{aREMBO}   [aREMBO] {active embedding Bayesian optimization}
\newacro{BO}       [BO] {Bayesian optimization}
\newacro{BBO}      [BBO] {Batch Bayesian optimization}
\newacro{ARD}      [ARD] {automatic relevance determination}
\newacro{RBF}      [RBF] {Radial basis function kernel}

%% file: neurips_2023.bbl
\begin{thebibliography}{10}
\providecommand{\url}[1]{#1}
\csname url@samestyle\endcsname
\providecommand{\newblock}{\relax}
\providecommand{\bibinfo}[2]{#2}
\providecommand{\BIBentrySTDinterwordspacing}{\spaceskip=0pt\relax}
\providecommand{\BIBentryALTinterwordstretchfactor}{4}
\providecommand{\BIBentryALTinterwordspacing}{\spaceskip=\fontdimen2\font plus
\BIBentryALTinterwordstretchfactor\fontdimen3\font minus \fontdimen4\font\relax}
\providecommand{\BIBforeignlanguage}[2]{{%
\expandafter\ifx\csname l@#1\endcsname\relax
\typeout{** WARNING: IEEEtran.bst: No hyphenation pattern has been}%
\typeout{** loaded for the language `#1'. Using the pattern for}%
\typeout{** the default language instead.}%
\else
\language=\csname l@#1\endcsname
\fi
#2}}
\providecommand{\BIBdecl}{\relax}
\BIBdecl

\bibitem{wang2020multi}
Y.~Wang, P.~D. Franzon, D.~Smart, and B.~Swahn, ``Multi-fidelity surrogate-based optimization for electromagnetic simulation acceleration,'' \emph{ACM Transactions on Design Automation of Electronic Systems (TODAES)}, vol.~25, no.~5, pp. 1--21, 2020.

\bibitem{torun2018global}
H.~M. Torun, M.~Swaminathan, A.~K. Davis, and M.~L.~F. Bellaredj, ``A global bayesian optimization algorithm and its application to integrated system design,'' \emph{IEEE Transactions on Very Large Scale Integration (VLSI) Systems}, vol.~26, no.~4, pp. 792--802, 2018.

\bibitem{lyu2018batch}
W.~Lyu, F.~Yang, C.~Yan, D.~Zhou, and X.~Zeng, ``Batch bayesian optimization via multi-objective acquisition ensemble for automated analog circuit design,'' in \emph{International conference on machine learning}.\hskip 1em plus 0.5em minus 0.4em\relax PMLR, 2018, pp. 3306--3314.

\bibitem{bergstra2013hyperopt}
J.~Bergstra, D.~Yamins, D.~D. Cox \emph{et~al.}, ``Hyperopt: A python library for optimizing the hyperparameters of machine learning algorithms,'' in \emph{Proceedings of the 12th Python in science conference}, vol.~13.\hskip 1em plus 0.5em minus 0.4em\relax Citeseer, 2013, p.~20.

\bibitem{snoek2012practical}
J.~Snoek, H.~Larochelle, and R.~P. Adams, ``Practical bayesian optimization of machine learning algorithms,'' \emph{Advances in neural information processing systems}, vol.~25, 2012.

\bibitem{brochu2010bayesian}
E.~Brochu, T.~Brochu, and N.~De~Freitas, ``A bayesian interactive optimization approach to procedural animation design,'' in \emph{Proceedings of the 2010 ACM SIGGRAPH/Eurographics Symposium on Computer Animation}, 2010, pp. 103--112.

\bibitem{wang2013bayesian}
Z.~Wang, M.~Zoghi, F.~Hutter, D.~Matheson, and N.~De~Freitas, ``Bayesian optimization in high dimensions via random embeddings,'' in \emph{Twenty-Third international joint conference on artificial intelligence}, 2013.

\bibitem{eriksson2019scaling}
D.~Eriksson, M.~Pearce, J.~Gardner, R.~D. Turner, and M.~Poloczek, ``Scalable global optimization via local bayesian optimization,'' \emph{Advances in neural information processing systems}, vol.~32, 2019.

\bibitem{binois2020practical}
M.~Binois, R.~B. Gramacy, and M.~Ludkovski, ``Practical heteroscedastic gaussian process modeling for large simulation experiments,'' \emph{Journal of Computational and Graphical Statistics}, vol.~27, no.~4, pp. 808--821, 2018.

\bibitem{lu2024high}
J.~Lu and R.~J. Zhu, ``High dimensional bayesian optimization via condensing-expansion projection,'' \emph{arXiv preprint arXiv:2408.04860}, 2024.

\bibitem{malu2021bayesian}
M.~Malu, G.~Dasarathy, and A.~Spanias, ``Bayesian optimization in high-dimensional spaces: A brief survey,'' in \emph{2021 12th International Conference on Information, Intelligence, Systems \& Applications (IISA)}.\hskip 1em plus 0.5em minus 0.4em\relax IEEE, 2021, pp. 1--8.

\bibitem{daxberger2020mixed}
E.~Daxberger, A.~Makarova, M.~Turchetta, and A.~Krause, ``Mixed-variable bayesian optimization,'' \emph{arXiv preprint arXiv:1907.01329}, 2019.

\bibitem{swersky2013multi}
K.~Swersky, J.~Snoek, and R.~P. Adams, ``Multi-task bayesian optimization,'' \emph{Advances in neural information processing systems}, vol.~26, 2013.

\bibitem{wang2016bayesian}
Z.~Wang, F.~Hutter, M.~Zoghi, D.~Matheson, and N.~De~Feitas, ``Bayesian optimization in a billion dimensions via random embeddings,'' \emph{Journal of Artificial Intelligence Research}, vol.~55, pp. 361--387, 2016.

\bibitem{letham2020re}
B.~Letham, R.~Calandra, A.~Rai, and E.~Bakshy, ``Re-examining linear embeddings for high-dimensional bayesian optimization,'' \emph{arXiv preprint arXiv:2001.11659}, 2020.

\bibitem{sankar2006smoothed}
A.~Sankar, D.~A. Spielman, and S.-H. Teng, ``Smoothed analysis of the condition numbers and growth factors of matrices,'' \emph{SIAM Journal on Matrix Analysis and Applications}, vol.~28, no.~2, pp. 446--476, 2006.

\end{thebibliography}
